\documentclass{article} 
\usepackage{iclr2027_conference,times}

\usepackage{amsmath,amsfonts,bm}

\def\eqref#1{equation~\ref{#1}}

\def\1{\bm{1}}

\DeclareMathAlphabet{\mathsfit}{\encodingdefault}{\sfdefault}{m}{sl}
\SetMathAlphabet{\mathsfit}{bold}{\encodingdefault}{\sfdefault}{bx}{n}

\usepackage{booktabs}
\usepackage{multirow}
\usepackage{hyperref}
\hypersetup{hidelinks}
\usepackage{url}
\usepackage{graphicx}
\usepackage{subcaption}
\usepackage{algorithm}
\usepackage{algpseudocode}
\usepackage{amssymb}
\usepackage{xcolor}

\definecolor{mohaPromotedFill}{HTML}{E6F1E9}
\definecolor{mohaRejectedFill}{HTML}{F9EADD}
\definecolor{mohaStatusInk}{HTML}{24313B}

\newcommand{\mohaStatusCell}[2]{%
  \begingroup
  \setlength{\fboxsep}{1.4pt}%
  \colorbox{#1}{%
    \makebox[3.7em][c]{%
      \color{mohaStatusInk}%
      \rule[-0.55ex]{0pt}{2.4ex}%
      \vphantom{$+21.9^{(1)}$}%
      #2%
    }%
  }%
  \endgroup
}

\newcommand{\mohaPromoted}[1]{%
  \mohaStatusCell{mohaPromotedFill}{#1}%
}

\newcommand{\mohaRejected}[1]{%
  \mohaStatusCell{mohaRejectedFill}{#1}%
}

\newcommand{\mohaNotTested}{%
  \textcolor{black!45}{\textemdash}%
}

\newcommand{\mohaStatusKey}[2]{%
  \colorbox{#1}{\rule{0pt}{0.6em}\hspace{0.55em}}\,#2%
}
\definecolor{acceptgreen}{HTML}{167C68}
\definecolor{rejectred}{HTML}{B64952}
\definecolor{neutralgray}{HTML}{6B7280}

\title{MoHA: Model-Conditioned Harness Adaptation for Long-Video Agents}

\author{%
    \textbf{Jianghan Chao\textsuperscript{1,2},
    Wenhui Tan\textsuperscript{1},
    Hang Su\textsuperscript{2},
    Qiuyu Tian\textsuperscript{2},
    Tanxiao Li\textsuperscript{1},
    Bingzi Zhang\textsuperscript{1}}\\
    \textbf{Zequn Liu\textsuperscript{2},
    Ruihua Song\textsuperscript{1},
    Yingce Xia\textsuperscript{2}}\\[2pt]
  \textsuperscript{1}\,Gaoling School of Artificial Intelligence,
  Renmin University of China\\
  \textsuperscript{2}\,Zhongguancun Academy\\
  \texttt{chaojh@ruc.edu.cn}
}

\iclrfinalcopy
\begin{document}

\maketitle

\begin{abstract}
Long-video understanding increasingly relies on agentic systems that actively acquire evidence rather than process fixed video representations.
Their execution harnesses must coordinate language-level planning with video-level observation and evidence acquisition, yet are typically fixed across model stacks.
We therefore introduce MoHA, a framework for model-conditioned harness adaptation that adapts a common initial execution harness to the specific planner and observer models it hosts.
MoHA treats the harness itself as a model-conditioned optimization object: under the same task distribution and initialization, different planner--observer stacks can benefit from different executable harnesses.
Starting from a common initial harness, MoHA analyzes failed calibration traces to diagnose model--harness mismatches and adapts both planner-side reasoning support, such as overview, memory, and verification, and observer-side capability assignments, including OCR and ASR specialists.
After structural adaptation, MoHA separately calibrates observer execution configurations for the hosted model stack.
Across three frozen planner stacks, MoHA improves Video-MME Full accuracy by 6.2--27.1 percentage points and Video-Holmes accuracy by 3.1--20.1 points over the fixed initial harness. Cross-model transfer further shows that the adapted harnesses generalize only partially to other models, highlighting the harness as a model-conditioned optimization layer for long-video understanding.
Code is available at \url{https://github.com/roverx12345/MoHA}.
\end{abstract}

\section{Introduction}
\label{sec:intro}

Long-video understanding remains challenging not only because videos contain far more visual content than can be processed densely, but also because useful evidence is sparse and unevenly distributed over time.
Recent methods therefore increasingly move beyond processing a fixed video representation and instead actively decide what evidence to acquire and how to acquire it.
Some methods adaptively select frames or regions during reasoning~\citep{framethinker2025,videobrain2026,videoo32026}, while external video agents use retrieval, memory, navigation, and perceptual tools to inspect videos on demand~\citep{videoagent2024,lenswalk2026,videoarm2026,videoseek2026,retoolvideo2026}.
As a result, long-video performance increasingly depends not only on the underlying models, but also on the execution harness that coordinates reasoning, evidence acquisition, and perception.

\begin{figure*}[t]
    \centering
    \includegraphics[width=0.95\linewidth]{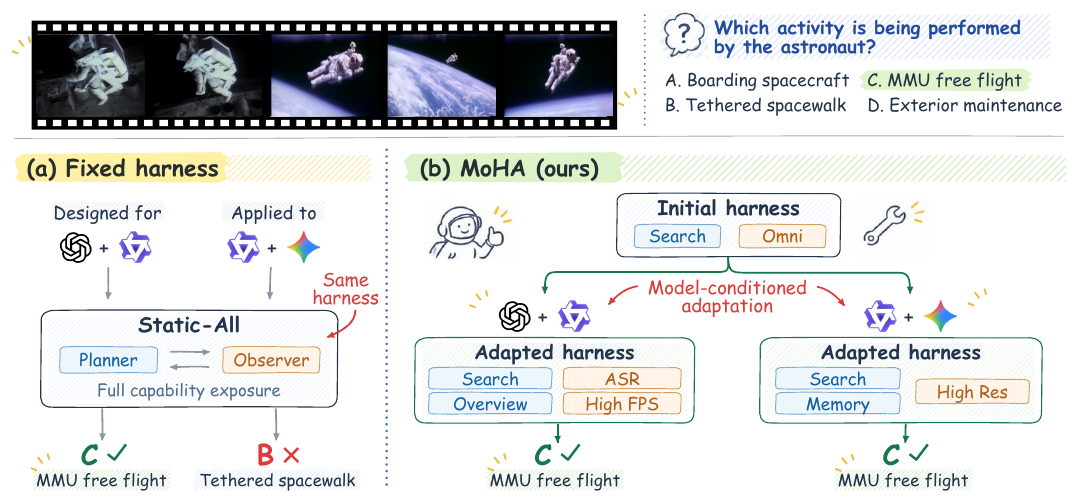}
    \caption{
    \textbf{Motivation for model-conditioned harness adaptation.}
    (a) A fixed harness is shared across planner--observer stacks despite their different failure patterns; exposing all available capabilities does not necessarily resolve this mismatch.
    (b) Starting from the same lightweight harness, MoHA adapts planner-facing modules, observer assignments, and execution configurations to each hosted stack.
    }
    \label{fig:teaser}
\end{figure*}

Existing agentic long-video systems typically pair a model stack with a prescribed execution pipeline that specifies the planner's reasoning support, the routing of perceptual requests, and the processing of video evidence.
Recent work has shown that the computation surrounding frozen models can itself be optimized~\citep{selfharness2026}.
However, improving a harness for one model stack does not establish whether the resulting design is equally well suited to others.
A natural alternative is to expose all available capabilities to every stack, but this still leaves the harness model-agnostic and does not account for which forms of reasoning support or perception each stack can use effectively.
Planners may differ in how effectively they search for, retain, integrate, and verify evidence, while observers may differ in which perceptual requests they handle reliably and which execution configurations they benefit from.
These differences motivate adapting the harness not only to the task, but also to the reasoning and perceptual capabilities of the models it hosts (see Figure~\ref{fig:teaser}).

We therefore introduce \textbf{MoHA}, a framework for \emph{model-conditioned harness adaptation} in long-video agents.
Starting from a common lightweight harness, MoHA adapts the harness to each planner--observer stack in two stages, while keeping model weights and the underlying runtime fixed.
First, it uses failed calibration trajectories to diagnose recurrent model--harness mismatches and propose bounded, localized structural interventions.
These interventions modify the reasoning support of the planner, such as overview, memory, or verification, or route suitable observation requests to OCR or ASR specialists.
Each candidate intervention is retained only when it improves held-out validation performance under matched execution budgets. After structural adaptation, MoHA freezes the resulting harness structure and separately calibrates observer configurations, including temporal sampling rate and visual input scale.
This separation distinguishes which capabilities are exposed to the model stack from how its perceptual capabilities are executed.

We evaluate MoHA across three frozen planner stacks on Video-MME and Video-Holmes. Starting from the same lightweight harness, MoHA consistently improves all three stacks, with gains of 6.2--27.1 percentage points on Video-MME Full and 3.1--20.1 points on Video-Holmes.
Model-conditioned adaptation is particularly effective on long videos.
Our analysis further reveals that the preferred harness changes meaningfully across models.
For example, adding verification improves held-out performance for Qwen3.5-27B but hurts the other two planners, while harnesses adapted to one planner transfer only partially to others.
Together, these results show that effective long-video harnesses are shaped not only by the task and available capabilities, but also by the models they host.

Our contributions are summarized as follows:

\noindent(1) \textbf{Model-conditioned harness adaptation for long-video agents.}
We formulate the execution harness as a model-conditioned optimization layer around a frozen planner--observer stack, allowing different model stacks to derive different validated harnesses from the same initialization and adaptation distribution.

\noindent(2) \textbf{Trace-driven structural adaptation with separate observer calibration.}
We introduce a bounded procedure that uses failed calibration trajectories to propose localized changes to planner-side reasoning support and observer capability assignments, retains them through held-out validation, and separately calibrates observer execution configurations after the harness structure is frozen.

\noindent(3) \textbf{Empirical gains and model-dependent harness behavior.}
MoHA consistently improves the evaluated planner stacks over the common initial harness on both Video-MME and Video-Holmes.
We further show that failure patterns, intervention effects, and cross-model transfer behavior differ across models, demonstrating that effective long-video harnesses depend on the models they host.

\section{Related Work}
\label{sec:related}

\paragraph{Long-Video Reasoning.}
Long-video understanding has progressed from extending or compressing visual
context~\citep{longva2024,longvila2025,longvu2025} to stronger temporal and
spatiotemporal reasoning~\citep{videor12025,videochatr12025,longvilar12025}.
Recent work further learns adaptive evidence acquisition through iterative
frame selection or tool-conditioned reasoning~\citep{framethinker2025,
videobrain2026,videoo32026}.
Across these approaches, the adaptation target ranges from visual
representations and context construction to learned reasoning and evidence
acquisition policies.
They primarily improve the video model or its learned inference behavior,
whereas MoHA keeps the hosted planner--observer stack frozen and adapts the
execution system surrounding it.

\paragraph{Agentic Video Understanding.}
A complementary line organizes long-video inference as an active interaction
process, using retrieval and adaptive observation~\citep{videoagent2024,
videotree2025,vca2024,lenswalk2026,videoseek2026,avp2025}, hierarchical or
long-term memory~\citep{videoarm2026,videolucy2025,m3agent2025,worldmm2026},
and specialized tools or reasoning roles~\citep{dvd2025,videothinker2026,
videomind2025,symphony2026,videoseal2026}.
These systems demonstrate that evidence acquisition and execution structure are
critical for long-video reasoning, but typically instantiate a particular agent
pipeline or interaction policy.
More recently, the video-agent system itself has become an optimization target:
MetaVideoAgent evolves modular video agents for a target video
distribution~\citep{metavid2026}, while VideoHarness-RSI searches executable
context-construction programs around a frozen vision-language
model~\citep{videoharnessrsi2026}.
MoHA builds on this direction, but conditions harness adaptation on the
\emph{hosted model stack}, and explicitly studies adaptation and transfer across
different planners and observers.

\paragraph{Agent Harness Optimization.}
Beyond video, earlier agent systems showed that interfaces, memory, and
execution orchestration can substantially shape fixed-model
behavior~\citep{react2023,memgpt2023,sweagent2024,autogen2023}.
More recent work directly optimizes the computation surrounding language models (LMs), including modular programs built around LMs, agent functions, prompts, and compound AI systems~\citep{dspy2024,agentoptimizer2024,textgrad2025,gepa2026}.
Other approaches externalize or search execution logic and harness
structure~\citep{nlah2026,metaharness2026}, while Self-Harness uses execution
traces to identify model-specific weaknesses and validate localized harness
modifications~\citep{selfharness2026}.
These methods mainly optimize language-side execution components.
MoHA extends this model-conditioned perspective to long-video agents, where
adaptation additionally spans learned perceptual observers and their execution
configurations.

\begin{figure*}[t]
    \centering
    \includegraphics[width=\linewidth]{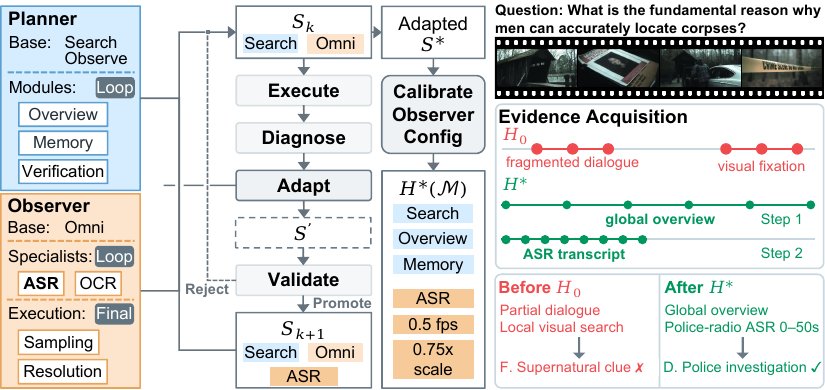}
    \caption{
    \textbf{Overview of MoHA.}
    Starting from a shared initial harness \(H_0=(S_0,c_0)\), MoHA collects calibration traces, diagnoses recurrent model--harness mismatches, and iteratively proposes and validates localized structural interventions.
    Planner-facing modules and observer capability assignments are adapted while \(c_0\) remains fixed, producing the final adapted structure \(S^*\).
    MoHA then freezes \(S^*\) and calibrates the observer execution configuration, yielding the final model-conditioned harness \(H^*(\mathcal M)=(S^*,c^*)\).
    Loop denotes components adapted during structural adaptation; Final denotes post-structural execution calibration.
    Right: an example in which global overview and ASR expose previously missing evidence and correct the initial answer.
    }
    \label{fig:method}
\end{figure*}


\section{Preliminaries}
\label{sec:preliminaries}

Long-video agents combine reasoning models with mechanisms for retrieval,
perception, memory, and iterative evidence acquisition~\citep{videoagent2024,videoarm2026,lenswalk2026,videoseek2026}.
As these systems actively interact with video content, their behavior depends
not only on the underlying models but also on the execution harness that
coordinates reasoning and perception.
We therefore treat the harness as an explicit design object.

Let
\(
\mathcal M=(M_p,\mathcal O)
\)
denote a frozen planner--observer model stack, where \(M_p\) is the planner
responsible for evidence acquisition and final answering, and
\(\mathcal O\) is the pool of available perceptual backends.
The pool contains a general-purpose multimodal observer and may additionally
contain specialist backends.

We write the adaptable harness configuration as \(H=(S,c)\), where \(S\)
denotes the \emph{harness structure} and \(c\) denotes the observer execution
configuration.
The Harness Runtime \(\mathcal R\) is fixed throughout adaptation and is
therefore omitted from the variable harness notation.
For a video question \(x\), the resulting agent produces
\(
\hat y=\mathrm{Run}(\mathcal M,H,\mathcal R;x)
\).

\paragraph{System overview.}
At a high level, a long-video agent alternates between planning and evidence
acquisition.
The planner uses the harness to locate relevant video regions and request
evidence from selected regions.
The structure \(S\) determines the planner support that is available and which
perceptual capabilities can handle observation requests, while the configuration
\(c\) determines how a selected backend processes the video.

\paragraph{Planner-facing structure.}
The planner-facing portion of \(S\) specifies the harness-level support
available to the planner during iterative evidence acquisition and reasoning.
Such support may affect how the agent locates relevant content, organizes and
retains acquired evidence, or re-examines evidence before producing a final
answer.
At minimum, a long-video harness must allow the planner to identify relevant
video content and request targeted perceptual evidence; richer harnesses may
provide additional support around these basic interactions.

We represent an observation request by a video region \(W_t\) and a typed
observation goal \(g_t\), specifying respectively \emph{where} to inspect and
\emph{what evidence} is sought.
The planner determines \((W_t,g_t)\), but does not choose the perceptual backend
or low-level execution settings.

\paragraph{Observer-facing structure and execution.}
The harness uses the observation goal to route each request to a perceptual
backend available under \(S\).
The configuration \(c\) then determines how the selected backend processes the
requested video region, including settings such as temporal sampling and visual
input scale.
This separates \emph{which perceptual capabilities are exposed and assigned}
from \emph{how the selected backends are executed}.

\paragraph{Harness Runtime.}
The fixed Harness Runtime \(\mathcal R\) executes planner operations, dispatches
observation requests, maintains intermediate state and execution traces, and
enforces inference budgets.


\section{Model-Conditioned Harness Adaptation}
\label{sec:adaptive-harness}

Given a frozen model stack \(\mathcal M\), MoHA adapts a common initial harness
using a calibration set \(\mathcal D_{\mathrm{cal}}\) and a video-disjoint
held-out validation set \(\mathcal D_{\mathrm{val}}\), neither of which contains
benchmark test examples.
All model stacks start from the same initial harness \(H_0=(S_0,c_0)\),
where \(S_0\) is the initial harness structure and \(c_0\) is the common
observer execution configuration.

MoHA separates adaptation into two stages.
It first adapts the structure \(S\), which determines the planner-facing support
and observer-facing capabilities available to the hosted model stack, while
keeping \(c_0\) fixed.
After the structure is frozen, MoHA separately calibrates the observer execution
configuration \(c\).
Model weights and the Harness Runtime remain fixed throughout.
Figure~\ref{fig:method} summarizes the overall procedure.
For readability, we omit the dependence on \(\mathcal M\) from intermediate
notation and make it explicit again for the final harness.

\subsection{Structural Harness Adaptation}
\label{sec:structural-adaptation}

The structural stage determines which additional planner-facing support and observer-facing capabilities are useful for the hosted model stack.
Starting from \(S_0\), MoHA uses failed calibration trajectories to propose localized structural changes and retains a change only when it improves held-out validation performance.

\paragraph{Adaptation space.}
The initial structure \(S_0\) instantiates the minimal planner--harness
interaction described in Section~\ref{sec:preliminaries} with two fixed
operations.
\texttt{search} identifies candidate temporal regions relevant to the current
question, while \texttt{observe} requests targeted evidence from a selected
region.
No additional planner-facing support is initially active, and all observation
requests are handled by a general-purpose multimodal observer.

MoHA draws structural harness updates from a small, fixed intervention catalog
\(\mathcal I\):
\begin{equation}
\mathcal{I} = \{
\texttt{overview},
\texttt{memory},
\texttt{verification},
\texttt{OCR},
\texttt{ASR}
\}.
\end{equation}
Planner-facing interventions may activate \texttt{overview}, \texttt{memory},
or \texttt{verification}, while observer-facing interventions may route eligible
text- or speech-oriented observation requests to OCR or ASR specialists.
Each intervention modifies a single previously inactive component.
The base \texttt{search} and \texttt{observe} operations remain unchanged, and
structural updates are additive in our implementation.

These components address different recurring failure patterns.
For example, a planner that repeatedly enters localized search without first
establishing useful global context may benefit from \texttt{overview}, whereas
one that acquires useful evidence but later loses access to it may instead
benefit from \texttt{memory}.
MoHA does not redesign these components themselves; it determines which of them
should be exposed to a particular model stack.

At round \(k\), let
\(
\mathcal I_k\subseteq\mathcal I
\)
denote the interventions still available under the current harness state,
excluding components that are already active or unavailable.

\paragraph{Trace-driven proposal.}
At round \(k\), MoHA executes the current structure \(S_k\) with the fixed
configuration \(c_0\) on \(\mathcal D_{\mathrm{cal}}\) and collects failed
execution trajectories.
For each failed trajectory \(\tau_i\), a structured LLM judge analyzes the
question, acquired evidence, interaction context, execution trajectory, and
final prediction.
It assigns a failure type from a bounded taxonomy
(see Table~\ref{tab:failure-taxonomy}) and identifies the trajectory steps
supporting its diagnosis.

For non-observer failures, the judge may recommend at most one currently
available intervention when the diagnosis and recorded trace support a clear
structural change; otherwise, it makes no recommendation.
Failure types do not map deterministically to interventions: the same failure
category may lead to different recommendations depending on the trajectory.

Observer failures undergo an additional diagnostic probe before MoHA considers
adding an OCR or ASR specialist.
For example, if the planner localizes a relevant segment and requests speech
evidence but the general-purpose observer fails to recover the dialogue, MoHA
does not immediately add ASR.
Instead, it repeats the same request on the same region with the same observer
under alternative execution settings (e.g., a different frame rate or visual input scale).
If any alternative recovers the requested evidence, the failure can be resolved
without introducing a specialist, so no specialist recommendation is made.
If none of the feasible alternatives succeeds, the corresponding specialist may
then be recommended.
These alternative settings are used only for diagnosis, while \(c_0\) remains
fixed during structural adaptation.
Full details of the failure taxonomy and diagnostic probe are provided in
Appendix~\ref{app:adaptation-details}.

\paragraph{Aggregation, validation, and update.}
MoHA aggregates trajectory-level recommendations using equal votes.
If round \(k\) contains \(M_k\) failed trajectories, the support for candidate
\(a\) is
\begin{equation}
    s_k(a)
    =
    \sum_{i=1}^{M_k}
    \mathbf{1}[a_i=a],
    \qquad
    a\in\mathcal I_k.
    \label{eq:recommendation-score}
\end{equation}
Only valid intervention recommendations are counted.
Candidates with nonzero support are ranked by \(s_k(a)\), with a fixed
deterministic tie-breaking rule.
The aggregated failure statistics are used only for analysis and do not affect the ranking.

The ranked candidates are then evaluated sequentially on
\(\mathcal D_{\mathrm{val}}\),
up to a fixed per-round candidate limit \(L\).
Let \(U_{\mathrm{val}}(S;c)\) denote validation accuracy for structure \(S\)
under observer configuration \(c\).
For candidate \(a\), \(S_k\oplus a\) denotes the structure obtained by applying
that localized intervention while leaving all other components unchanged.
A candidate is promoted only if
\begin{equation}
    U_{\mathrm{val}}(S_k\oplus a;c_0)
    -
    U_{\mathrm{val}}(S_k;c_0)
    >
    \epsilon.
    \label{eq:validation-criterion}
\end{equation}
Thus, diagnosis determines which modifications are worth testing, not which
modification is accepted.
For example, even if \texttt{overview} is repeatedly recommended across failed
trajectories, it is rejected if its held-out gain does not exceed
\(\epsilon\).

The first candidate satisfying the criterion becomes \(S_{k+1}\), so at most
one structural intervention is promoted per round.
Rejected candidates are not reconsidered under the same structure.
Structural adaptation stops when the round budget is exhausted, no admissible
intervention receives support, or no supported candidate passes held-out
validation.
We denote the resulting structure by \(S^*\).

\subsection{Observer Configuration Calibration}
\label{sec:observer-config-calibration}

Structural adaptation determines \emph{which} capabilities the harness should
contain; this second stage determines \emph{how} the resulting observer system
should be executed.
With \(S^*\) fixed, MoHA evaluates a bounded candidate set
\(\mathcal C\) of observer execution configurations, varying settings such as
temporal sampling and visual input scale under matched sensory budgets.
Planner-side modules and observer capability assignments remain unchanged
throughout this stage.

MoHA selects
\begin{equation}
    c^*
    =
    \arg\max_{c\in\mathcal C}
    U_{\mathrm{val}}(S^*;c),
    \label{eq:observer-config-selection}
\end{equation}
and forms the final model-conditioned harness
\begin{equation}
    H^*(\mathcal M)
    =
    (S^*,c^*).
    \label{eq:final-harness}
\end{equation}
Together with the fixed Harness Runtime, \(H^*(\mathcal M)\) is frozen before
benchmark evaluation.
Further adaptation details are provided in
Appendix~\ref{app:adaptation-details}.


\section{Experiments}
\label{sec:experiments}

We evaluate MoHA across multiple planner--observer stacks on long-video understanding
benchmarks.
Beyond overall performance, we further analyze how the adapted harnesses behave
across models and which components of the adaptation process contribute to the gains.


\subsection{Experimental Setup}
\label{sec:exp-setup}

\paragraph{Datasets and evaluation protocol.}
We use the Video-Holmes training split~\citep{videoholmes2025} as an independent adaptation source, with a fixed video-disjoint split of 64 calibration and 32 held-out validation examples shared across all model stacks.
No benchmark test examples are used during adaptation.
After adaptation, each harness is frozen and evaluated on
Video-MME-v1~\citep{videomme2025} and the Video-Holmes test set.
Controlled transfer and ablation studies use Video-MME-v1 Long.
Video-MME-v1 is evaluated without provided subtitles; audio remains available to the harness for speech evidence acquisition.

\paragraph{Model stacks.}
We evaluate frozen planner--observer stacks
$\mathcal{M}=(M_p,\mathcal{O})$, all starting from the same harness $H_0$.
For planner-conditioned adaptation, we compare GPT-5.5~\citep{gpt552026},
Qwen3.5-27B, and Qwen3.5-9B~\citep{qwen352026} with
Qwen3-Omni-30B-A3B~\citep{qwen3omni2025} as the observer.
For observer-conditioned transfer, we fix GPT-5.5 and compare
Qwen3-Omni-30B-A3B with Qwen2.5-Omni-7B~\citep{qwen25omni2025}.
All non-varied components and execution budgets are held fixed.

\paragraph{Baselines.}
\emph{Static-All} exposes all available planner-side modules and specialist observer assignments without model-conditioned structural selection, while retaining observer-configuration calibration.
\emph{Initial} uses the common harness $H_0$ without adaptation.
For \emph{Shared}, all model stacks jointly adapt a single harness.
At each adaptation round, the stacks run under the same current harness, and
their calibration evidence is pooled to propose harness updates, which are
then evaluated using validation evidence pooled across the stacks.
This produces one adapted harness that is shared by all model stacks.
\emph{MoHA} instead adapts a separate harness $H^*(\mathcal{M})$ for each
hosted stack.

\paragraph{Implementation and budgets.}
All controlled comparisons use at most 16 planner calls and the same
per-observation limit of 128 sampled frames.
For Video-MME-v1, we additionally report the average number of sampled video
frames consumed per question as a measure of realized visual cost.
Additional adaptation settings, observer configuration candidates, and implementation details are provided in Appendix~\ref{app:experimental-details}.
A trajectory that does not emit a legal answer option within the planner-call budget is counted as incorrect.
The 128-frame limit applies to an individual observation request; it is not a per-question total-frame limit for the full agent execution.
Direct Omni baselines use one request per question and are reported in the upper block of Table~\ref{tab:overall-controlled}.


\subsection{Overall Performance}
\label{sec:overall}

\begin{table*}[t]
    \centering
    \caption{
    \textbf{Overall performance.} Initial uses the common harness \(H_0\); Shared adapts one harness from
    evidence pooled across model stacks; MoHA adapts a separate harness for
    each hosted stack.
    Short, Medium, Long, Full, and Video-Holmes report accuracy (\%).
    \emph{Frames/Q} is the average number of sampled video frames consumed
    per question on Video-MME-v1.
    Bold indicates the best result within each planner block.
    The upper block reports direct-answer observer baselines.
    }
    \label{tab:overall-controlled}
    \setlength{\tabcolsep}{4.4pt}
    \renewcommand{\arraystretch}{1.05}

    \begin{tabular}{llccccc c}
        \toprule
        \multirow{2}{*}{Model}
        & \multirow{2}{*}{Harness}
        & \multicolumn{5}{c}{Video-MME-v1}
        & \multirow{2}{*}{Video-Holmes} \\
        \cmidrule(lr){3-7}
        &
        & Short
        & Medium
        & Long
        & Full
        & Frames/Q $\downarrow$
        & \\
        \midrule
        \multicolumn{8}{l}{\textit{Direct-observer baselines}} \\
        \midrule
        Qwen2.5-Omni & -- & 72.8 & 63.6 & 45.7 & 60.7 & 128.0 & 40.1 \\
        Qwen3-Omni & -- & 79.4 & 66.8 & 49.0 & 65.1 & 128.0 & 51.2 \\
        \midrule
        \multicolumn{8}{l}{\textit{Planner--observer harness comparisons}} \\
        \midrule

        \multirow{4}{*}{GPT-5.5}
        & Static-All
        & 71.0 & 66.9
        & 75.7 & 71.2
        & 126.2 & 42.6 \\
        & Initial $H_0$
        & 74.3 & 74.4 & 66.2 & 71.7 & 174.3 & 47.7 \\
        & Shared
        & \textbf{81.4} & \textbf{78.0} & 72.1 & 77.2
        & 164.9 & \textbf{52.1} \\
        & \textbf{MoHA}
        & 78.8 & 77.7 & \textbf{77.3} & \textbf{77.9}
        & \textbf{98.2} & 51.9 \\

        \midrule

        \multirow{4}{*}{Qwen3.5-27B}
        & Static-All
        & 67.6 & 50.8
        & 37.6 & 52.0
        & 105.6 & 23.2 \\
        & Initial $H_0$
        & 72.4 & 65.6 & 53.9 & 64.0 & 106.1 & 17.9 \\
        & Shared
        & \textbf{78.9} & 59.1 & 53.7 & 63.9
        & 85.6 & 23.2 \\
        & \textbf{MoHA}
        & 78.3 & \textbf{72.2} & \textbf{65.2} & \textbf{71.9}
        & \textbf{60.6} & \textbf{38.0} \\

        \midrule

        \multirow{4}{*}{Qwen3.5-9B}
        & Static-All
        & 59.3 & 48.6
        & 44.3 & 50.7
        & 54.1 & 14.8 \\
        & Initial $H_0$
        & 43.2 & 31.2 & 22.8 & 32.4 & 33.7 & 26.3 \\
        & Shared
        & 67.2 & \textbf{62.0} & 42.1 & 57.1
        & 53.8 & \textbf{30.9} \\
        & \textbf{MoHA}
        & \textbf{74.6} & 52.7 & \textbf{51.1} & \textbf{59.5}
        & \textbf{28.0} & 29.4 \\

        \bottomrule
    \end{tabular}
\end{table*}

Table~\ref{tab:overall-controlled} compares the common initial harness,
the jointly adapted Shared harness, and model-conditioned MoHA.
On Video-MME-v1 Full, MoHA improves over $H_0$ by 6.2, 7.9, and
27.1 percentage points for GPT-5.5, Qwen3.5-27B, and Qwen3.5-9B,
respectively.
The gains are larger on the Long subset, reaching 11.1, 11.3, and
28.3 points.

Shared adaptation substantially improves some stacks, indicating that some harness modifications transfer across models.
However, it does not consistently recover the benefit of stack-specific
adaptation.
On Video-MME-v1 Long, MoHA exceeds Shared by 5.2 points for GPT-5.5,
11.5 for Qwen3.5-27B, and 9.0 for Qwen3.5-9B.
Shared can nevertheless match or outperform MoHA on individual shorter-video
subsets, while the advantage of model-specific adaptation is most consistent on the Long subset.
Moreover, simply exposing all capabilities (\emph{Static-All}) consistently underperforms MoHA across all three stacks—and even degrades below $H_0$ on Video-Holmes for every planner—confirming that unconditional capability addition is suboptimal.

MoHA also improves over $H_0$ on the Video-Holmes test set
for all three stacks, by 4.2, 20.1, and 3.1 points for GPT-5.5,
Qwen3.5-27B, and Qwen3.5-9B, respectively.
Notably, these gains do not come from consuming more sampled video
frames. On Video-MME-v1, MoHA reduces Frames/Q by 43.7\%, 42.9\%,
and 16.9\% relative to $H_0$ for the three planners, respectively,
while improving Full accuracy in all cases.
Compared with Shared, MoHA likewise achieves higher Full accuracy
with fewer sampled frames for every stack.
Together, these results show that model-conditioned adaptation can improve task performance while reducing realized visual sampling cost.

The upper block of Table~\ref{tab:overall-controlled} complements the matched Initial--Shared--MoHA comparison with direct-observer baselines.
While direct Qwen3-Omni is competitive on short videos (79.4\% on Short), its accuracy drops sharply to 49.0\% on Video-MME-v1 Long, where GPT-5.5 MoHA reaches 77.3\% (and 77.9\% vs.\ 65.1\% on Full).
On Video-Holmes, direct Qwen3-Omni outperforms the two Qwen planner stacks, largely because 99.56\% of Video-Holmes videos are no longer than 300 seconds and thus fit within the direct observer's single-pass audio budget
(Appendix~\ref{app:system-baseline-protocols}).


\subsection{Cross-Model Harness Transfer}
\label{sec:cross-model-transfer}

The overall gains do not by themselves establish that the adapted harness
depends on the hosted model: different adaptation runs could simply discover
equivalent generally useful configurations.
We therefore transfer each frozen $H^*$ to another model without further
adaptation.
For planner-conditioned transfer, we hold the general-purpose observer and
specialist pool fixed and replace only the planner.
For observer-conditioned transfer, we fix GPT-5.5 and replace only the
general-purpose Omni observer.

\noindent
\begin{minipage}{\linewidth}

\captionsetup{hypcap=false}
\captionof{table}{
\textbf{Cross-model harness transfer on Video-MME-v1 Long.}
Rows denote the model used at evaluation time, while each
\(H^*_{\cdot}\) column denotes a harness adapted for the model indicated
by its subscript and transferred without further adaptation.
Diagonal entries therefore correspond to matched model--harness pairs,
whereas off-diagonal entries measure cross-model transfer.
Initial denotes the shared starting harness \(H_0\).
Bold indicates the matched adapted harness in each row.
}
\label{tab:model-conditioned-transfer}

\vspace{2pt}

\normalsize

\renewcommand{\arraystretch}{1.06}

\begin{minipage}[t]{0.535\linewidth}
\vspace{0pt}
\centering

\textbf{(a) Planner-conditioned transfer}

\vspace{2pt}

{\setlength{\tabcolsep}{5.5pt}
\begin{tabular}{@{}lcccc@{}}
\toprule
Planner
& Initial
& $H^*_{\mathrm{GPT}}$
& $H^*_{\mathrm{Q27B}}$
& $H^*_{\mathrm{Q9B}}$ \\
\midrule

GPT-5.5
& 66.2
& \textbf{77.3}
& 71.6
& 67.7 \\

Qwen3.5-27B
& 53.9
& 50.4
& \textbf{65.2}
& 57.0 \\

Qwen3.5-9B
& 22.8
& 41.9
& 35.5
& \textbf{51.1} \\

\bottomrule
\end{tabular}
}

\end{minipage}%
\hspace{0.01\linewidth}%
\begin{minipage}[t]{0.445\linewidth}
\vspace{0pt}
\centering

\textbf{(b) Observer-conditioned transfer}

\vspace{2pt}

{\renewcommand{\arraystretch}{1.22}
\setlength{\tabcolsep}{4.5pt}
\begin{tabular}{@{}lccc@{}}
\toprule
Observer
& Initial
& $H^*_{\mathrm{Q2.5O}}$
& $H^*_{\mathrm{Q3O}}$ \\
\midrule

Qwen2.5-Omni
& 50.4
& \textbf{57.8}
& 54.7 \\

Qwen3-Omni
& 66.2
& 67.3
& \textbf{77.3} \\

\bottomrule
\end{tabular}
}

\end{minipage}

\end{minipage}

\vspace{0.6em}

Table~\ref{tab:model-conditioned-transfer} shows that the matched harness performs best in every planner-transfer comparison.
For example, Qwen3.5-27B reaches 65.2 with $H^*_{\mathrm{Q27B}}$, compared with 50.4 under $H^*_{\mathrm{GPT}}$ and 57.0 under $H^*_{\mathrm{Q9B}}$.
Observer-conditioned transfer shows the same pattern: the matched harness performs best for both Omni observers, indicating that model dependence is not limited to the planner.

At the same time, transfer is only partially successful and can be
asymmetric.
For example, $H^*_{\mathrm{GPT}}$ raises Qwen3.5-9B from 22.8 under
$H_0$ to 41.9, yet lowers Qwen3.5-27B from 53.9 to 50.4.
Thus, an adapted harness is not simply better or worse in isolation:
its effect depends on the model that hosts it.


\subsection{Ablations}
\label{sec:ablation}

We study whether MoHA's gains can be reproduced by exposing all available capabilities, by removing individual adaptation surfaces, or by replacing diagnosis-guided proposals with random proposals.

The adaptation-surface ablations use the GPT-5.5/Qwen3-Omni stack on Video-MME-v1 Long. Static-All exposes all available planner-side modules and specialist observer assignments without model-conditioned structural selection, while retaining observer-configuration calibration.
The random-proposal controls use the Qwen3.5-9B splits under the same intervention library, adaptation data, validation procedure, and observer-configuration budget.

\noindent
\begin{minipage}[t]{0.52\textwidth}
\captionsetup{hypcap=false}
\captionof{table}{Ablation of MoHA adaptation surfaces on Video-MME-v1 Long.}
\label{tab:ablation}
\normalsize
\setlength{\tabcolsep}{2.5pt}
\renewcommand{\arraystretch}{1.08}

\centering
\begin{tabular}{@{}lr@{}}
\toprule
Variant & Accuracy \tabularnewline
\midrule
Initial $H_0$ & 66.2 \tabularnewline
Static-All & 75.7 \tabularnewline
\midrule
w/o planner-side adaptation & 69.1 \tabularnewline
w/o specialist adaptation & 72.1 \tabularnewline
w/o observer config calibration & 75.9 \tabularnewline
\midrule
\textbf{MoHA} & \textbf{77.3} \tabularnewline
\bottomrule
\end{tabular}
\end{minipage}
\hfill
\begin{minipage}[t]{0.43\textwidth}
\captionsetup{hypcap=false}
\captionof{table}{Random-proposal controls on Qwen3.5-9B.}
\label{tab:random-proposals}
\normalsize
\setlength{\tabcolsep}{2.5pt}
\renewcommand{\arraystretch}{1.08}

\centering
\begin{tabular}{@{}lr@{}}
\toprule
Run & Accuracy \tabularnewline
\midrule
MoHA & 51.1 \tabularnewline
\midrule
Random split 1 & 46.1 \tabularnewline
Random split 2 & 45.6 \tabularnewline
Random split 3 & 44.1 \tabularnewline
\bottomrule
\end{tabular}
\end{minipage}

\vspace{4pt}

For the GPT-5.5/Qwen3-Omni stack, Static-All remains 1.6 points below MoHA (Table~\ref{tab:ablation}).
Removing planner-side adaptation causes the largest decrease of 8.2 points, followed by removing specialist adaptation and observer-configuration calibration, which decrease accuracy by 5.2 and 1.4 points, respectively.
These results indicate that exposing more capabilities without model-conditioned structural selection does not fully recover the benefit of MoHA.

The random-proposal controls achieve lower Long accuracy than MoHA in all three Qwen3.5-9B runs (Table~\ref{tab:random-proposals}).
This result supports the value of diagnosis-guided proposal selection over random proposal selection under the same adaptation setting.


\subsection{Analysis of Model-Dependent Behavior}
\label{sec:analysis}

\begin{figure*}[t]
    \centering
    \begin{subfigure}[c]{0.5\linewidth}
        \centering
        \includegraphics[width=\linewidth]{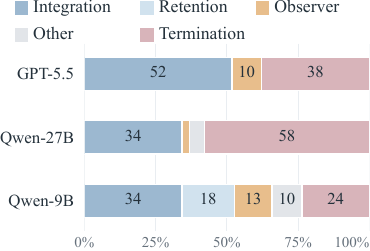}
        \caption{Failure profiles under $H_0$.}
        \label{fig:analysis-failure}
    \end{subfigure}%
    \hspace{0.012\linewidth}%
    \begin{subfigure}[c]{0.485\linewidth}
        \centering
        \small
        \renewcommand{\arraystretch}{1.08}
        \setlength{\tabcolsep}{2pt}

        \begin{tabular*}{\linewidth}{
            @{\extracolsep{\fill}}
            lccc
            @{}
        }
            \toprule
            &
            \textbf{GPT-5.5}
            & \textbf{Qwen-27B}
            & \textbf{Qwen-9B} \\
            \midrule

            Overview
            & \mohaRejected{$-6.2\,\times$}
            & \mohaPromoted{$+21.9^{(1)}$}
            & \mohaPromoted{$+6.2^{(2)}$} \\
            \addlinespace[1.2pt]

            Memory
            & \mohaPromoted{$+4.7^{(1)}$}
            & \mohaRejected{$+1.6\,\times$}
            & \mohaPromoted{$+7.8^{(1)}$} \\
            \addlinespace[1.2pt]

            Verify
            & \mohaRejected{$-21.9\,\times$}
            & \mohaPromoted{$+3.1^{(2)}$}
            & \mohaRejected{$-26.6\,\times$} \\
            \addlinespace[1.2pt]

            ASR
            & \mohaPromoted{$+3.1^{(2)}$}
            & \mohaNotTested
            & \mohaPromoted{$+3.1^{(3)}$} \\
            \addlinespace[1.2pt]

            OCR
            & \mohaNotTested
            & \mohaNotTested
            & \mohaRejected{$-10.9\,\times$} \\
            \addlinespace[1.2pt]

            \midrule

            FPS
            & $0.5$
            & $0.5$
            & $2.0$ \\

            Scale
            & $0.50$
            & $0.75$
            & $0.75$ \\

            \bottomrule
        \end{tabular*}

        \par\vspace{4pt}
        {\fontsize{8.5}{10.5}\selectfont
         \setlength{\fboxsep}{1pt}%
         \mohaStatusKey{mohaPromotedFill}{Promoted}%
         \hspace{0.7em}%
         \mohaStatusKey{mohaRejectedFill}{Rejected}%
         \hspace{0.7em}%
         \mohaNotTested\ Not tested
         \par
        }

        \caption{Intervention effects and selected harnesses.}
        \label{fig:analysis-intervention}
    \end{subfigure}

    \caption{
    \textbf{Analysis of model-conditioned harness adaptation.}
    (a) Failure profiles over unsuccessful calibration trajectories under
    the common initial harness $H_0$.
    (b) Validation effects of tested interventions and final harness choices.
    Values are accuracy changes in percentage points; superscripts indicate
    promotion order and $\times$ denotes rejected candidates.
    Final rows report the selected FPS and input scale.
    }
    \label{fig:moha-analysis}
\end{figure*}

We finally examine how model dependence appears within the adaptation process.
Figure~\ref{fig:moha-analysis} compares failure patterns under the shared initial harness with the validation effects and final outcomes of subsequent structural interventions.

\paragraph{Different models fail differently under the same initial harness.}
Figure~\ref{fig:analysis-failure} considers only unsuccessful trajectories produced under $H_0$, before any model-specific adaptation.
For visualization, trajectories that do not produce a valid final answer due to step-budget termination, final-output parsing failure, a missing recognizable answer, or abstention are grouped as \emph{Termination};
infrequent semantic diagnosis categories are merged into \emph{Other}.
Termination accounts for 38\%, 58\%, and 24\% of unsuccessful trajectories for GPT-5.5, Qwen3.5-27B, and Qwen3.5-9B, respectively.
The semantic composition of the remaining failures also differs across planners, showing that the same initial harness exposes different weaknesses for different hosted models.

\paragraph{The same intervention can reverse its effect across models.}
Figure~\ref{fig:analysis-intervention} combines the held-out validation effect of each tested intervention with the resulting promotion path and final observer configuration.
Each value reports the accuracy change relative to the current harness state at the round in which that candidate is evaluated.
Overview, for example, changes validation accuracy by $-6.2$ points for GPT-5.5, but by $+21.9$ for Qwen3.5-27B and $+6.2$ for Qwen3.5-9B.
Verification instead improves Qwen3.5-27B by $+3.1$ points while reducing GPT-5.5 and Qwen3.5-9B by $21.9$ and $26.6$ points, respectively.
Model dependence therefore appears not only in effect magnitude but even in effect direction: a capability that benefits one hosted model can be harmful for another.
This motivates held-out validation rather than unconditional capability addition.
A positive change can still be rejected when it does not exceed the promotion margin, as with Memory for Qwen3.5-27B ($+1.6$ points).

The resulting harnesses consequently differ in both structure and perceptual execution.
GPT-5.5 promotes Memory followed by ASR and selects $0.5$ FPS with input scale $0.5$.
Qwen3.5-27B promotes Overview followed by Verification and selects $0.5$ FPS with scale $0.75$.
Qwen3.5-9B promotes Memory, Overview, and ASR and selects $2.0$ FPS with scale $0.75$.
Together with the cross-model transfer results, these differences show that model conditioning affects both which capabilities are exposed and how the resulting perceptual system is executed.


\section{Conclusions and Future Work}

We introduced MoHA, a framework for adapting the execution harness of a long-video agent to its frozen planner--observer stack.
Across three planner stacks, MoHA improves over the common initial harness on both Video-MME and Video-Holmes, while cross-model transfer and intervention analyses show that effective harness configurations depend on the models they host.
These results support treating the execution harness as a model-conditioned adaptation target rather than a fixed component of long-video agents.
Future work could explore automatic harness-component discovery, joint model--harness co-adaptation, and broader evaluation across models, domains, and multimodal agent tasks.

\clearpage

\subsection*{AI Use Statement}

Generative AI tools were used as assistive tools during this work, primarily for improving the clarity and presentation of author-developed ideas, providing feedback on experimental design, assisting with code editing and debugging, summarizing relevant literature, and polishing the manuscript and scientific figures. They were also used to assist in organizing and interpreting experimental results, with all analyses and conclusions independently reviewed and verified by the authors. The central research question, core methodological ideas, and main scientific contributions of this work were developed by the authors rather than generated by AI tools. All experimental decisions, implementations, reported results, and final claims were reviewed and validated by the authors, who take full responsibility for the content of the paper.

\subsection*{Reproducibility Statement}

We provide the information necessary to reproduce our experiments throughout the main paper and appendix. The main paper specifies the MoHA adaptation procedure, model stacks, datasets, evaluation protocol, and inference budgets. The appendix further documents the harness components, calibration and validation procedure, prompts, observer configuration calibration, implementation details, and additional experimental settings and analyses. We additionally provide the code repository linked in the abstract, including scripts for running calibration, harness adaptation, and evaluation.

\bibliography{iclr2027_conference}
\bibliographystyle{iclr2027_conference}


\appendix


\section{Harness and Adaptation Details}
\label{app:harness-details}
\label{app:adaptation-details}

We provide implementation details omitted from the main text.
Following Section~\ref{sec:preliminaries}, the adaptable harness is
\(
H=(S,c)
\),
where \(S\) denotes the executable harness structure and \(c\) the observer
execution configuration.
The Harness Runtime \(\mathcal R\) is fixed throughout adaptation and is
therefore omitted from the variable harness notation.

\subsection{Harness Components and Observer Execution}
\label{app:harness-components}
\label{app:observer-execution}
\label{app:component-config}

MoHA operates over a small, fixed component library (Table~\ref{tab:executable-components}).
The base \texttt{search} and \texttt{observe} operations are always available;
structural adaptation only activates optional planner support or changes
observer assignments.

\begin{table*}[t]
    \centering
    \caption{
    Executable components in the MoHA harness.
    Base operations remain available throughout adaptation, while optional
    planner modules and specialist assignments may be activated.
    }
    \label{tab:executable-components}
    \small
    \setlength{\tabcolsep}{5pt}
    \begin{tabular}{
        p{0.14\textwidth}
        p{0.18\textwidth}
        p{0.29\textwidth}
        p{0.29\textwidth}}
        \toprule
        Component & Role & Behavior & Adaptable effect \\
        \midrule

        \texttt{search}
        & Base planner operation
        & Retrieves candidate temporal regions
        & Always available \\

        \texttt{observe}
        & Base planner operation
        & Acquires evidence from region \(W_t\) for typed goal \(g_t\)
        & Always available \\

        \texttt{overview}
        & Planner support
        & Provides coarse global video orientation
        & May be activated \\

        \texttt{memory}
        & Planner support
        & Preserves bounded information across reasoning steps
        & May be activated \\

        \texttt{verification}
        & Planner support
        & Supports explicit re-examination of accumulated evidence
        & May be activated \\

        Omni observer
        & Default observer
        & Handles general observation requests
        & Remains the fallback backend \\

        OCR
        & Specialist observer
        & Handles typed text-evidence requests
        & Eligible requests may be reassigned from Omni \\

        ASR
        & Specialist observer
        & Handles typed speech-evidence requests
        & Eligible requests may be reassigned from Omni \\
        \bottomrule
    \end{tabular}
\end{table*}

For an observation step, the planner selects a video support \(W_t\) and a
typed semantic goal \(g_t\).
The active structure \(S\) determines the observer \(o_t\): text or speech
requests are routed to an active OCR or ASR specialist, respectively, while
all remaining requests use the general-purpose observer.
The selected backend is executed under configuration \(c\):
\begin{equation}
    z_t =
    \mathrm{Observe}
    \left(
        W_t,
        g_t,
        o_t;
        c
    \right).
\end{equation}
Thus, the planner determines \emph{where} to inspect and \emph{what} evidence
is needed, while the harness determines which available backend executes the
request and how it is executed.

MoHA separates structural and execution-level adaptation.
During structural adaptation, \(c=c_0\) is fixed and only \(S\) may change.
After the final structure \(S^*\) is frozen, observer configuration calibration
selects \(c^*\) from a bounded candidate set.
Temporary execution changes used by the diagnostic probe below do not update
\(c_0\).

\subsection{Failure Diagnosis}
\label{app:failure-diagnosis}

At structural adaptation round \(k\), MoHA executes the current structure
\(S_k\) with fixed configuration \(c_0\) on the calibration set.
For each failed trajectory, a structured LLM judge attributes the failure using
the question, acquired evidence, execution trace, and final prediction.
Deterministic checks validate output structure, referenced trajectory steps,
capability--request consistency, and candidate legality.
Runtime or infrastructure failures are handled separately.

The judge uses the bounded taxonomy in
Table~\ref{tab:failure-taxonomy}.
The taxonomy is diagnostic rather than prescriptive: a failure category does
not deterministically map to an intervention.

\begin{table*}[t]
    \centering
    \caption{Failure taxonomy used for calibration-trace diagnosis.}
    \label{tab:failure-taxonomy}
    \small
    \setlength{\tabcolsep}{6pt}
    \begin{tabular}{p{0.19\textwidth}p{0.73\textwidth}}
        \toprule
        Failure & Interpretation \\
        \midrule

        Orientation
        & The planner fails to establish a useful region of the video before
        detailed evidence acquisition. \\

        Retrieval
        & Relevant temporal candidates are not retrieved despite a plausible
        evidence need. \\

        Candidate selection
        & A useful candidate is retrieved but not selected or followed up. \\

        Goal specification
        & The selected region is plausible, but the observation request asks
        for insufficient or mismatched evidence. \\

        Observer
        & The region and request are plausible, but the assigned observer does
        not recover evidence supporting the requested goal. \\

        Evidence retention
        & Useful acquired evidence is no longer available to later reasoning
        steps. \\

        Evidence integration
        & Relevant evidence remains available but is combined or used
        incorrectly. \\

        Verification
        & The trajectory contains evidence motivating re-examination, but the
        planner commits without resolving the remaining conflict. \\

        Unresolved
        & The trace does not support a reliable attribution to the categories
        above. \\
        \bottomrule
    \end{tabular}
\end{table*}

Non-\texttt{unresolved} diagnoses must reference at least one recorded
trajectory step.
OCR or ASR capability attribution is valid only for an
\texttt{observer} failure containing a corresponding typed text or speech
request.

\subsection{Observer Specialist Gating Probe}
\label{app:observer-probe}

An \texttt{observer} diagnosis does not immediately justify adding OCR or ASR.
Instead, MoHA first tests whether the cited general-purpose-observer failure can
be rescued by changing only its execution configuration.

The probe re-executes the cited observation while keeping the video asset,
source window, selected region, typed observation goal, and general-purpose
observer fixed.
Planner decisions, search trajectories, and final answer generation are not
replayed, and the observer call does not receive the answer label.
The probe varies only execution fields, using candidate values from
\[
\{0.5,1.0,2.0\}\ \text{FPS},
\qquad
\{0.5,0.75,1.0\}\ \text{visual scale},
\]
and
\(
\{\text{temporal},\text{spatial},\text{balanced}\}
\)
execution priority, excluding no-op, infeasible, or media-equivalent
configurations.
These temporary executions do not modify \(c_0\).

A structured probe judge evaluates whether the baseline and alternative
observations support the same requested goal.
An \emph{execution rescue} occurs when the baseline fails but at least one
alternative succeeds; in this case the trace is not treated as sufficient
evidence for introducing a specialist.
A \emph{no-rescue} result requires all feasible tested alternatives to fail with
valid judgments.
Other cases are treated as inconclusive.

Only a no-rescue result makes a matching specialist admissible:
OCR additionally requires a typed text request and OCR attribution, while ASR
requires a typed speech request and ASR attribution.
The specialist must still be recommended at the trajectory level and pass the
held-out validation gate described below.

\subsection{Structural Adaptation and Promotion}
\label{app:intervention-catalog}
\label{app:recommendation-ranking}
\label{app:validation-promotion}
\label{app:stopping}

Structural adaptation uses the fixed intervention catalog
\[
\mathcal I =
\{
\texttt{overview},
\texttt{memory},
\texttt{verification},
\texttt{OCR},
\texttt{ASR}
\}.
\]
Updates are additive: once activated, a module or specialist assignment remains
active in subsequent rounds.

Each failed calibration trajectory contributes at most one currently admissible
intervention recommendation, or makes no recommendation.
Invalid, unresolved, null, or illegal proposals contribute no vote.
Valid proposals receive equal weight and are ranked by support count, using the
candidate identifier as a deterministic tie-break.
There is no learned global selector after these trajectory-level
recommendations.

For candidate \(a\), let \(S_k\oplus a\) denote the structure obtained by
applying only that intervention to \(S_k\).
The current and candidate structures are evaluated on the same held-out
validation examples using the same frozen model stack, fixed configuration
\(c_0\), and matched execution and sensory budgets.
A candidate is promoted only when
\begin{equation}
    U_{\mathrm{val}}(S_k\oplus a;c_0)
    -
    U_{\mathrm{val}}(S_k;c_0)
    >
    \epsilon.
\end{equation}
Candidates are tested in ranked order up to the configured per-round limit, and
the first candidate passing the gate becomes \(S_{k+1}\).
Structural adaptation stops when the round budget is exhausted, no admissible
candidate receives support, or none of the tested candidates passes validation.

\subsection{Observer Configuration Calibration}
\label{app:observer-config-details}

After structural adaptation terminates, the resulting structure \(S^*\) is
frozen.
MoHA then selects the observer execution configuration on the held-out
validation set:
\begin{equation}
    c^*
    =
    \arg\max_{c\in\mathcal C}
    U_{\mathrm{val}}(S^*;c).
\end{equation}
No planner module or observer assignment changes during this stage.
The final harness is
\begin{equation}
    H^*(\mathcal M)=(S^*,c^*),
\end{equation}
which is frozen before benchmark evaluation.

In our experiments, \(\mathcal C\) is the Cartesian product of
\(\{0.5,1.0,2.0\}\) FPS and
\(\{0.5,0.75,1.0\}\) visual input scales with balanced execution priority,
under the same frame and visual-token ceilings used by the structural
comparisons.
Unlike the specialist gating probe, which asks whether one failed observation
can be rescued, this stage selects a single execution configuration for the
already frozen structure.

\subsection{Full Adaptation Algorithm}
\label{app:full-algorithm}

Algorithm~\ref{alg:full-calibration} summarizes the complete procedure.

\begin{algorithm}[t]
\caption{\textsc{Model-Conditioned Harness Adaptation}}
\label{alg:full-calibration}
\footnotesize
\begin{algorithmic}[1]

\Require frozen model stack $\mathcal M$; shared initial harness
$H_0=(S_0,c_0)$; calibration set $\mathcal D_{\mathrm{cal}}$;
validation set $\mathcal D_{\mathrm{val}}$; intervention catalog $\mathcal I$;
configuration set $\mathcal C$; execution budget $B$;
promotion margin $\epsilon$; maximum structural rounds $K$;
candidate evaluation limit $L$

\Ensure frozen model-conditioned harness $H^*(\mathcal M)$

\State $S\gets S_0$

\For{$k=0,\ldots,K-1$}

    \State $\mathcal T
    \gets
    \Call{CollectTraces}
    {\mathcal M,S,c_0,\mathcal D_{\mathrm{cal}},B}$

    \State $\mathcal I_k
    \gets
    \Call{Admissible}{\mathcal I,S}$

    \If{$\mathcal I_k=\emptyset$}
        \State \textbf{break}
    \EndIf

    \State $\mathcal P\gets[]$

    \ForAll{failed trajectories $\tau_i\in\mathcal T$}

        \State $d_i\gets\Call{Diagnose}{\tau_i,\mathcal I_k}$

        \If{$d_i.\mathrm{failure}=\texttt{observer}$}
            \State $p_i
            \gets
            \Call{ProbeObserver}{\tau_i,d_i,c_0}$
            \State $a_i
            \gets
            \Call{Recommend}{\tau_i,d_i,p_i,\mathcal I_k}$
        \Else
            \State $a_i\gets\Call{Recommendation}{d_i}$
        \EndIf

        \If{$a_i\neq\varnothing$}
            \State $\mathcal P\gets\operatorname{append}(\mathcal P,a_i)$
        \EndIf

    \EndFor

    \State $\mathcal Q\gets\Call{RankBySupport}{\mathcal P}$

    \If{$\mathcal Q=\emptyset$}
        \State \textbf{break}
    \EndIf

    \State $\mathrm{promoted}\gets\mathrm{false}$

    \For{the first $L=2$ candidates $a \in \mathcal Q$ in ranked order}

        \If{
        $\Call{ValAcc}{S\oplus a,c_0}
        -
        \Call{ValAcc}{S,c_0}
        >
        \epsilon$
        }
            \State $S\gets S\oplus a$
            \State $\mathrm{promoted}\gets\mathrm{true}$
            \State \textbf{break}
        \EndIf

    \EndFor

    \If{$\neg\mathrm{promoted}$}
        \State \textbf{break}
    \EndIf

\EndFor

\State $S^*\gets S$

\State $c^*
\gets
\Call{CalibrateObserverConfig}
{\mathcal M,S^*,\mathcal C,\mathcal D_{\mathrm{val}},B}$

\State $H^*(\mathcal M)\gets(S^*,c^*)$

\State \Return $\Call{Freeze}{H^*(\mathcal M)}$

\end{algorithmic}
\end{algorithm}


\section{Experimental and Implementation Details}
\label{app:experimental-details}

\subsection{Calibration and Validation Split}
\label{app:data-split}

Adaptation data are partitioned by video identity before harness optimization.
We use 64 calibration examples for trace collection, diagnosis, probing, and
intervention recommendation, and 32 held-out validation examples for structural
promotion and post-structural observer configuration selection.
No benchmark test example is used during adaptation.

\subsection{Adaptation Hyperparameters}
\label{app:hyperparameters}

Table~\ref{tab:adaptation-hyperparameters} summarizes the fixed
hyperparameters used across adaptation runs. Structural adaptation uses a
common observer configuration and a bounded number of rounds and candidate
evaluations; after the structure is frozen, observer calibration searches the
listed sampling-rate and visual-scale configurations.

\begin{table}[t]
    \centering
    \caption{MoHA adaptation hyperparameters.}
    \label{tab:adaptation-hyperparameters}
    \small
    \setlength{\tabcolsep}{5pt}
    \begin{tabular}{lc}
        \toprule
        Hyperparameter & Value \\
        \midrule
        Maximum structural rounds \(K\) & 6 \\
        Candidate limit \(L\) per round & 2 \\
        Promotion margin \(\epsilon\) & 0.02 \\
        Calibration examples & 64 \\
        Validation examples & 32 \\
        Validation repeats & 2 \\
        Structural-stage target FPS & 1.0 \\
        Structural-stage visual scale & 1.0 \\
        Structural-stage priority & balanced \\
        Frame cap & 128 \\
        Observer sampling candidates
        & \(\{0.5,1.0,2.0\}\) FPS \\
        Observer scale candidates
        & \(\{0.5,0.75,1.0\}\) \\
        \bottomrule
    \end{tabular}
\end{table}

\subsection{Model and Runtime Configuration}
\label{app:model-config}

All model weights remain frozen throughout adaptation.
The released configuration specifies the exact planner and observer snapshots,
specialist backends, maximum planner steps, observation and sensory budgets,
video decoding procedure, diagnosis and probe judge models, and API or local
inference settings used for each reported condition.
Consequently, differences among Initial, Shared, MoHA, transfer, and ablation
conditions arise from the execution harness or its controlled removal rather
than model fine-tuning.

\subsection{Diagnostic Interface and Prompts}
\label{app:diagnostic-prompts}

The structured diagnosis interface records the failure category, supporting
trajectory-step identifiers, optional OCR/ASR capability attribution, and at
most one currently available intervention recommendation.
The implementation additionally records confidence and short textual
explanations for attribution, residual uncertainty, and recommendation or
abstention.

Deterministic validation checks the required field set, legal enum values,
confidence range, evidence-step references, capability--request consistency,
and candidate legality.
Non-\texttt{unresolved} diagnoses require recorded supporting steps; OCR
attribution requires a typed text request and ASR attribution requires a typed
speech request.
For the initial \texttt{observer} diagnosis, the intervention recommendation
must be null until the specialist gating probe is resolved.
Invalid structured outputs receive at most one format-repair attempt without
forcing a semantic attribution.

Failure diagnosis and specialist probing use frozen Claude Opus 4.8
(\texttt{claude-opus-4-8}), separately from the planner and observer models.
Both interfaces return structured JSON outputs with a maximum of 4,096
completion tokens. Invalid structured outputs receive at most one format-repair
attempt, and validation checks enforce the required schema, legal enum values,
evidence-step references, and candidate legality.

The supplementary implementation provides the exact diagnosis and probe
prompts, structured schemas, and validation logic.
Neither diagnosis nor probe components receive held-out validation labels or
benchmark test examples.

\subsection{Common-Reference Module Analysis}
\label{app:common-memory}

To separate intervention effects from the adaptation path, we evaluated the
same planner-side interventions from a common Memory reference structure.
The evaluation uses the same 32 video-disjoint validation questions, two
repeats per condition, fixed observer settings, and matched execution budgets.

\begin{table}[t]
    \centering
    \caption{Controlled module comparison from a common Memory reference.}
    \label{tab:common-memory}
    \small
    \setlength{\tabcolsep}{4pt}
    \begin{tabular}{lccc}
        \toprule
        Model stack
        & Memory
        & Memory + Overview
        & Memory + Verification \tabularnewline
        \midrule
        Qwen3.5-9B
        & 24/64
        & 28/64 (+6.25 pp)
        & 7/64 (-26.56 pp) \tabularnewline
        Qwen3.5-27B
        & 27/64
        & 33/64 (+9.38 pp)
        & 24/64 (-4.69 pp) \tabularnewline
        \bottomrule
    \end{tabular}
\end{table}

As shown in Table~\ref{tab:common-memory}, Overview improves both stacks from the common Memory reference, by
6.25 percentage points for Qwen3.5-9B and 9.38 points for Qwen3.5-27B.
Verification instead decreases accuracy for both stacks, with a much larger
drop for Qwen3.5-9B. In the audited traces, adding Verification increases
abstentions from 5/64 to 41/64 for 9B and from 1/64 to 13/64 for 27B,
suggesting that the additional checking step can interfere with final answer
completion, particularly for Qwen3.5-9B under the Memory reference structure.

\subsection{Adaptation-Split Sensitivity}
\label{app:split-sensitivity}

We reran the complete Qwen3.5-9B adaptation and evaluation procedure on three new random, video-disjoint calibration--validation splits.
Each split used 64 calibration questions and 32 validation questions evaluated twice.

\begin{table*}[t]
    \centering
    \caption{Adaptation-split sensitivity for Qwen3.5-9B.}
    \label{tab:split-sensitivity}
    \small
    \setlength{\tabcolsep}{4pt}
    \begin{tabular}{lcccccc}
        \toprule
        Split
        & Calibration
        & Validation
        & Selected modules
        & FPS
        & Scale
        & Long accuracy \tabularnewline
        \midrule
        S0
        & 64 & 32
        & Overview + Memory + ASR
        & 2.0 & 0.75 & 50.7\% \tabularnewline
        S1
        & 64 & 32
        & Overview + Memory
        & 2.0 & 0.75 & 50.4\% \tabularnewline
        S2
        & 64 & 32
        & Memory + ASR
        & 0.5 & 0.5 & 46.7\% \tabularnewline
        Original MoHA
        & 64 & 32
        & Overview + Memory + ASR
        & 2.0 & 0.75 & 51.1\% \tabularnewline
        \bottomrule
    \end{tabular}
\end{table*}

Table~\ref{tab:split-sensitivity} shows that Memory is selected in all three runs, Overview and ASR in two runs each, and Verification in none, indicating that while core preferences are consistent, the exact selected structure and accuracy can vary across adaptation splits.

\subsection{Fixed-Harness Repeat Evaluation}
\label{app:repeat-evaluation}

To assess evaluation stability, we repeated evaluation with the GPT-5.5
MoHA harness fixed on a 600-question subset (the first 200 questions each from Video-MME-v1 Short, Medium, and Long, denoted Short200, Medium200, and Long200). The repeated run follows the dataset order and uses the same questions as the original run.

\begin{table}[t]
    \centering
    \caption{Fixed-harness repeat evaluation on a 600-question Video-MME-v1 subset.}
    \label{tab:repeat-evaluation}
    \small
    \setlength{\tabcolsep}{4pt}
    \begin{tabular}{lcccc}
        \toprule
        Evaluation
        & Short200
        & Medium200
        & Long200
        & Full600 \tabularnewline
        \midrule
        Original run
        & 80.0\% & 79.5\% & 88.5\% & 82.67\% \tabularnewline
        Repeat run
        & 81.5\% & 78.5\% & 89.0\% & 83.00\% \tabularnewline
        Change
        & +1.5 pp & -1.0 pp & +0.5 pp & +0.33 pp \tabularnewline
        \bottomrule
    \end{tabular}
\end{table}

Table~\ref{tab:repeat-evaluation} shows stable aggregate performance across repeated runs on this subset.

\subsection{Answer Validity and Budget Sensitivity}
\label{app:validity-budget}

Our primary metric is end-to-end accuracy under a fixed budget of 16 planner
calls. A trajectory that does not produce a legal answer option within this
budget is counted as incorrect, encompassing budget termination, final-output
parsing failure, missing recognizable answers, and abstention.

Table~\ref{tab:unresolved-answers} reports the number of unresolved initial-harness outputs after terminal-answer rescoring.
On Video-Holmes, the final planner output of Qwen3.5-27B still contains an \texttt{observe} or \texttt{search} request in 1,117/1,837 trajectories (60.8\%).
When the trajectory reaches the 16-call limit, no subsequent planner call remains to integrate the requested evidence and submit an answer; if the final output also lacks an extractable answer option, the episode is scored as invalid, accounting for the 54.3\% invalid-output rate for Qwen3.5-27B on Video-Holmes.

\begin{table}[t]
    \centering
    \caption{Unresolved initial-harness outputs after terminal-answer rescoring.}
    \label{tab:unresolved-answers}
    \small
    \setlength{\tabcolsep}{5pt}
    \begin{tabular}{lcc}
        \toprule
        Planner
        & Video-MME Long
        & Video-Holmes \tabularnewline
        \midrule
        Qwen3.5-27B
        & 135/900
        & 997/1,837 \tabularnewline
        Qwen3.5-9B
        & 526/900
        & 170/1,837 \tabularnewline
        \bottomrule
    \end{tabular}
\end{table}

As a scoring sensitivity analysis, Table~\ref{tab:random-answer-sensitivity} reports the expected initial-harness accuracies when replacing only unresolved outputs with a uniformly random answer option.

\begin{table*}[t]
    \centering
    \caption{Expected initial-harness accuracy after random replacement of unresolved outputs.}
    \label{tab:random-answer-sensitivity}
    \small
    \setlength{\tabcolsep}{5pt}
    \begin{tabular}{lccccc}
        \toprule
        Planner
        & Short
        & Medium
        & Long
        & Full
        & Video-Holmes \tabularnewline
        \midrule
        GPT-5.5
        & 74.4
        & 74.5
        & 66.2
        & 71.7
        & 47.8 \tabularnewline
        Qwen3.5-27B
        & 73.1
        & 67.4
        & 57.6
        & 66.0
        & 28.4 \tabularnewline
        Qwen3.5-9B
        & 53.3
        & 44.3
        & 37.4
        & 45.0
        & 27.8 \tabularnewline
        \bottomrule
    \end{tabular}
\end{table*}

To examine budget sensitivity directly, we ran 16-to-32-call continuations for all 526 originally invalid
Qwen3.5-9B \(H_0\) trajectories on Video-MME-v1 Long (Table~\ref{tab:budget-continuation}). Among these
trajectories, 57/526 (10.8\%) produced valid answers, 34/526 (6.5\%)
produced correct answers, and 469/526 (89.2\%) remained invalid.
Thus, doubling the planner-call budget from 16 to 32 leaves most unresolved $H_0$ trajectories unresolved, highlighting the need for structural execution support rather than simply extending the call budget.

\begin{table}[t]
    \centering
    \caption{16-to-32-call continuation analysis on Video-MME-v1 Long.}
    \label{tab:budget-continuation}
    \small
    \setlength{\tabcolsep}{4pt}
    \begin{tabular}{lccc}
        \toprule
        Condition
        & Scope
        & Valid / correct
        & Invalid \tabularnewline
        \midrule
        16-to-32 continuation
        & 526 originally invalid trajectories
        & 57 valid / 34 correct
        & 469 \tabularnewline
        \bottomrule
    \end{tabular}
\end{table}

Table~\ref{tab:failure-aggregation} summarizes the failure-category breakdown across 105 unsuccessful initial-harness calibration trajectories using mutually exclusive reporting categories.
Trajectories without a valid final answer are assigned to Termination, even
when a semantic diagnosis identifies an underlying issue. These failure
categories characterize the initial harness and are distinct from intervention
recommendations, which are accepted only after held-out validation.

\begin{table}[t]
    \centering
    \caption{Failure-category aggregation for unsuccessful calibration trajectories.}
    \label{tab:failure-aggregation}
    \small
    \setlength{\tabcolsep}{3pt}
    \begin{tabular}{lcccccc}
        \toprule
        Model
        & Termination
        & Evidence integration
        & Evidence retention
        & Observer
        & Other
        & Total \tabularnewline
        \midrule
        GPT-5.5
        & 11
        & 15
        & 0
        & 3
        & 0
        & 29 \tabularnewline
        Qwen3.5-27B
        & 22
        & 13
        & 0
        & 1
        & 2
        & 38 \tabularnewline
        Qwen3.5-9B
        & 9
        & 13
        & 7
        & 5
        & 4
        & 38 \tabularnewline
        \bottomrule
    \end{tabular}
\end{table}

\subsection{Adaptation and Deployment Cost}
\label{app:adaptation-cost}

We complement Frames/Q with measurements of diagnosis, probing,
auxiliary-model calls, token usage, and audio processing, reporting
adaptation and deployment resources separately.

Tables~\ref{tab:adaptation-calls} and~\ref{tab:adaptation-resources} summarize logged execution calls and average per-episode resource usage during adaptation.
Table~\ref{tab:auxiliary-cost} reports auxiliary diagnosis judge and specialist-probe usage separately.
The validation totals include approximately 320 structural candidate-validation episodes for each planner, plus 576 episodes for the final nine-configuration observer sweep in each run; diagnosis counts include logged retries.

\begin{table*}[t]
    \centering
    \caption{Logged adaptation episodes and execution calls. Main-call totals are recorded by an independent runtime counter rather than recomputed as the sum of planner and observer calls.}
    \label{tab:adaptation-calls}
    \small
    \setlength{\tabcolsep}{4pt}
    \begin{tabular}{lcc}
        \toprule
        Planner
        & Main calls: total / planner / observer
        & Audio seconds / episode \tabularnewline
        \midrule
        GPT-5.5
        & 11,525 / 11,461 / 6,683
        & 153.8 \tabularnewline
        Qwen3.5-27B
        & 16,661 / 16,597 / 9,822
        & 152.1 \tabularnewline
        Qwen3.5-9B
        & 16,749 / 16,685 / 7,494
        & 63.5 \tabularnewline
        \bottomrule
    \end{tabular}
\end{table*}

\begin{table*}[t]
    \centering
    \caption{Average main-execution resources per adaptation episode.}
    \label{tab:adaptation-resources}
    \small
    \setlength{\tabcolsep}{5pt}
    \begin{tabular}{lcc}
        \toprule
        Planner
        & Main text tokens / episode
        & Video tokens / episode \tabularnewline
        \midrule
        GPT-5.5
        & 32.7K
        & 16.4K \tabularnewline
        Qwen3.5-27B
        & 43.6K
        & 18.8K \tabularnewline
        Qwen3.5-9B
        & 25.9K
        & 8.36K \tabularnewline
        \bottomrule
    \end{tabular}
\end{table*}

\begin{table*}[t]
    \centering
    \caption{Auxiliary diagnosis and specialist-probe usage during adaptation.}
    \label{tab:auxiliary-cost}
    \small
    \setlength{\tabcolsep}{4pt}
    \begin{tabular}{lcc}
        \toprule
        Planner
        & Diagnosis judge
        & Specialist probe \tabularnewline
        \midrule
        GPT-5.5
        & 148 calls; 62.5K tokens/call
        & 5 calls; 4.00K text + 1.65K video tokens/call \tabularnewline
        Qwen3.5-27B
        & 262 calls; 47.1K tokens/call
        & 44 calls; 3.54K text + 1.35K video tokens/call \tabularnewline
        Qwen3.5-9B
        & 279 calls; 55.1K tokens/call
        & 14 embedded calls; 2.90K text + 0.89K video tokens/call \tabularnewline
        \bottomrule
    \end{tabular}
\end{table*}

Table~\ref{tab:deployment-resources} reports per-question deployment resource usage during benchmark evaluation.
Random search has no diagnosis or specialist-probe cost by design, but still incurs candidate-validation and deployment costs.

\begin{table*}[t]
    \centering
    \caption{Per-question deployment resources during benchmark evaluation.}
    \label{tab:deployment-resources}
    \small
    \setlength{\tabcolsep}{4pt}
    \begin{tabular}{lccccc}
        \toprule
        Planner
        & Planner calls/Q
        & Observer calls/Q
        & Text tokens/Q
        & Visual tokens/Q
        & Audio sec/Q \tabularnewline
        \midrule
        GPT-5.5
        & 5.87 & 4.93 & 57.47K & 6.60K & 264.90 \tabularnewline
        Qwen3.5-27B
        & 5.71 & 3.45 & 38.80K & 23.21K & 111.39 \tabularnewline
        Qwen3.5-9B
        & 4.43 & 1.97 & 21.23K & 9.51K & 22.74 \tabularnewline
        \bottomrule
    \end{tabular}
\end{table*}

\subsection{Direct-Omni Baseline Protocol}
\label{app:system-baseline-protocols}

The Direct Omni baselines use one request per question without a
planner--observer tool loop. Each request uses at most 128 uniformly sampled
video frames and 300 seconds of audio. Table~\ref{tab:system-baseline-protocols}
summarizes their results across Video-MME-v1 and Video-Holmes.

\begin{table*}[t]
    \centering
    \caption{Direct-observer baseline protocol and results.}
    \label{tab:system-baseline-protocols}
    \small
    \setlength{\tabcolsep}{5pt}
    \begin{tabular}{llccccc}
        \toprule
        Model
        & Evaluation scope
        & Short
        & Medium
        & Long
        & Full
        & Video-Holmes \\
        \midrule
        Qwen2.5-Omni
        & One request/question
        & 72.8 & 63.6 & 45.7 & 60.7 & 40.1 \\
        Qwen3-Omni
        & One request/question
        & 79.4 & 66.8 & 49.0 & 65.1 & 51.2 \\
        \bottomrule
    \end{tabular}
\end{table*}

To contextualize the direct-answer comparison, Table~\ref{tab:video-duration}
summarizes video-duration distributions across the two benchmarks.
Video-Holmes videos have mean and median durations of 185 and 190 seconds, and
99.56\% of its questions correspond to videos no longer than 300 seconds.
By contrast, Video-MME-v1 Long videos have mean and median durations of 2,466
and 2,423 seconds, and 897 of 900 questions correspond to videos longer than
30 minutes. The shorter Video-Holmes videos make direct answering more
competitive, whereas the much longer Video-MME-v1 Long videos motivate targeted
observation and cross-segment evidence integration.

\begin{table}[t]
    \centering
    \caption{Video-duration statistics for the two benchmarks.}
    \label{tab:video-duration}
    \small
    \begin{tabular}{lccc}
        \toprule
        Benchmark
        & Mean (s)
        & Median (s)
        & Duration coverage \\
        \midrule
        Video-Holmes
        & 185 & 190
        & 99.56\% \(\leq\) 300 s \\
        Video-MME-v1 Long
        & 2,466 & 2,423
        & 897/900 \(>\) 30 min \\
        \bottomrule
    \end{tabular}
\end{table}

Thus, the 300-second audio budget covers nearly all Video-Holmes videos but
only a small part of a typical Video-MME-v1 Long video. Longer videos place
greater demands on targeted observation and evidence integration across
distant segments. Performance depends jointly on the planner, observer, and
harness: effective long-video reasoning requires the planner to invoke the
observer appropriately and integrate the resulting evidence through the harness.

\subsection{Extended Module Library}
\label{app:extended-library}

The main experiments use a single basic implementation for each module family.
As an initial probe into extending the module library, we conducted an extension run for Qwen3.5-9B with an additional basic Memory implementation available from the initial harness.
As shown in Table~\ref{tab:extended-library}, the adapted harness selects the basic Memory module and converges to a 2.0 FPS, 0.5-scale observation configuration.

\begin{table}[t]
    \centering
    \caption{Extended module-library run for Qwen3.5-9B.}
    \label{tab:extended-library}
    \small
    \setlength{\tabcolsep}{5pt}
    \begin{tabular}{lccc}
        \toprule
        Planner
        & Added module
        & FPS
        & Visual scale \tabularnewline
        \midrule
        Qwen3.5-9B
        & Memory
        & 2.0
        & 0.5 \tabularnewline
        \bottomrule
    \end{tabular}
\end{table}


\section{Relation to Prior Harness and Agent Adaptation}
\label{app:additional-comparisons}
\label{app:closest-methods}

Table~\ref{tab:harness-comparison} summarizes the closest lines of work along
the dimensions most relevant to MoHA.
The comparison is intended to clarify adaptation targets rather than rank
methods.

\begin{table*}[t]
    \centering
    \caption{
    Comparison with representative work that optimizes the execution system
    surrounding a frozen or hosted model.
    }
    \label{tab:harness-comparison}
    \small
    \setlength{\tabcolsep}{5pt}
    \begin{tabular}{
        p{0.18\textwidth}
        p{0.19\textwidth}
        p{0.25\textwidth}
        p{0.30\textwidth}}
        \toprule
        Method
        & Primary conditioning focus
        & Adapted object
        & Distinction relative to MoHA \\
        \midrule

        MetaVideoAgent~\citep{metavid2026}
        & Target video / task distribution
        & Modular video-agent design
        & Searches for an effective agent design for a target distribution;
        MoHA instead keeps the task distribution and initialization shared and
        asks how the harness changes with the hosted model stack. \\

        VideoHarness-RSI~\citep{videoharnessrsi2026}
        & Frozen video-model answering interface
        & Executable context-construction program
        & Optimizes the inference program around a fixed interface; MoHA
        separately exposes planner support, observer assignment, and observer
        execution configuration and studies how these change across model
        stacks. \\

        Self-Harness~\citep{selfharness2026}
        & Model-specific execution traces
        & Localized harness modifications
        & Shares the model-conditioned, trace-driven perspective; MoHA extends
        it to long-video agents with learned perceptual backends and explicit
        observer-side adaptation. \\

        Natural-Language Agent Harnesses /
        MetaHarness~\citep{nlah2026,metaharness2026}
        & General agent execution logic
        & Harness / execution program
        & Treats harness logic as an optimization object more broadly; MoHA
        uses a bounded component space to enable interpretable model-conditioned
        adaptation and controlled cross-model transfer analysis. \\

        \textbf{MoHA}
        & \textbf{Hosted planner--observer stack}
        & \textbf{Planner modules, observer assignment, and observer execution}
        & \textbf{Uses shared initialization and adaptation data across model
        stacks and explicitly evaluates whether adapted harnesses transfer
        across planners and observers.} \\
        \bottomrule
    \end{tabular}
\end{table*}


\section{Representative Long-Video Agent Results}
\label{app:video-agent-comparison}

For literature context, Table~\ref{tab:representative-video-agents} reports
published Video-MME v1 results from representative long-video agents and
tool-using systems.
These systems differ in backbone models, audiovisual access, training data,
retrieval procedures, and inference budgets, so the numbers should not be
interpreted as controlled comparisons.
Our primary controlled evidence is the matched Initial--Shared--MoHA evaluation
in the main text.

\begin{table}[t]
    \centering
    \caption{
        Representative reported Video-MME v1 results.
        Values are taken from the corresponding papers and are provided as
        literature context rather than a controlled leaderboard.
    }
    \label{tab:representative-video-agents}
    \small
    \setlength{\tabcolsep}{8pt}
    \renewcommand{\arraystretch}{1.08}
    \begin{tabular}{lcc}
        \toprule
        Method & Long & Full \\
        \midrule

        VideoAgent~\citep{videoagent2024}
            & 46.4 & -- \\

        VideoTree~\citep{videotree2025}
            & 54.2 & 60.6 \\

        VideoLucy~\citep{videolucy2025}
            & 66.8 & 72.5 \\

        AVP (Gemini 2.5 Pro)~\citep{avp2025}
            & 81.9 & 85.3 \\

        LensWalk (o3/GPT-4.1)~\citep{lenswalk2026}
            & 70.0 & -- \\

        VideoARM (o3/GPT-4.1)~\citep{videoarm2026}
            & 75.3 & 80.1 \\

        VideoSeek (GPT-5)~\citep{videoseek2026}
            & 70.1 & -- \\

        ReTool-Video-9B~\citep{retoolvideo2026}
            & -- & 76.6 \\

        \midrule

        \textbf{MoHA (GPT-5.5)}
            & \textbf{77.3} & \textbf{77.9} \\

        \textbf{MoHA (Qwen3.5-27B)}
            & \textbf{65.2} & \textbf{71.9} \\

        \textbf{MoHA (Qwen3.5-9B)}
            & \textbf{51.1} & \textbf{59.5} \\

        \bottomrule
    \end{tabular}
\end{table}


\section{Limitations}

MoHA is evaluated on a limited set of planner--observer stacks and long-video
benchmarks.
Although different stacks benefit from different adapted harnesses in our
experiments, broader evaluation across model families, video domains, and
multimodal tasks is needed to establish generality.
MoHA also searches within a predefined space of planner-facing modules and
observer capabilities rather than arbitrary harness programs or dynamically
created capabilities.

The validation set contains 32 distinct examples, each evaluated twice,
and is reused for both structural promotion and the final nine-way observer
configuration sweep. Split-sensitivity experiments (Appendix~\ref{app:split-sensitivity}) show that
the selected structure and final accuracy can vary across calibration--validation
splits. The fixed-harness repeat evaluation (Appendix~\ref{app:repeat-evaluation}) provides stability
evidence from two evaluations of the same GPT-5.5 configuration on a
600-question subset.

Adaptation also depends on calibration-trace diagnoses, which may be imperfect for ambiguous failures, and incurs additional inference cost before deployment.
Some experiments rely on externally served models that may exhibit run-to-run
stochasticity.
Finally, MoHA keeps planner and observer parameters frozen and adapts only the
surrounding execution harness; joint model--harness co-adaptation remains a
promising direction for future work.

\end{document}